\documentclass[sigconf]{acmart}
\usepackage{booktabs} 
\usepackage{url}
\usepackage{multirow}
\usepackage{mathtools}
\usepackage{amsthm}
\usepackage[utf8]{inputenc}
\PassOptionsToPackage{numbers, compress}{natbib}
\usepackage{natbib}
\usepackage{xr-hyper,refcount}
\usepackage{nicefrac}       
\usepackage{bbm}
\usepackage{mathtools}
\usepackage{cases}
\usepackage{comment}
\usepackage{subcaption}
\usepackage{algorithm}
\usepackage{hhline}
\usepackage{tabularx}
\usepackage{makecell}
\usepackage{color}
\usepackage{appendix}
\usepackage{xspace}
\usepackage{enumitem}
\usepackage{soul}
\usepackage[geometry]{ifsym}
\usepackage{pifont}
\usepackage{multirow}
\usepackage{microtype}

\newcommand{\red}{$R$}
\newcommand{\uone}{$U_1$}
\newcommand{\utwo}{$U_2$}
\newcommand{\syn}{$S$}

\usepackage{caption} 
\usepackage{amsmath,amsfonts,bm}

\def\eqref#1{equation~\ref{#1}}

\def\1{\bm{1}}

\DeclareMathAlphabet{\mathsfit}{\encodingdefault}{\sfdefault}{m}{sl}
\SetMathAlphabet{\mathsfit}{bold}{\encodingdefault}{\sfdefault}{bx}{n}

\DeclareMathOperator*{\argmax}{arg\,max}

\copyrightyear{2023}
\acmYear{2023}
\setcopyright{rightsretained}
\acmConference[ICMI '23]{International Conference on Multimodal Interaction}{October 9--13, 2023}{Paris, France}
\acmBooktitle{International Conference on Multimodal Interaction (ICMI '23), October 9--13, 2023, Paris, France}
\acmDOI{10.1145/3577190.3614151}
\acmISBN{979-8-4007-0055-2/23/10}

\begin{document}
\title{Multimodal Fusion Interactions: A Study of Human and Automatic Quantification}

\author{Paul Pu Liang, Yun Cheng, Ruslan Salakhutdinov, Louis-Philippe Morency}
\email{pliang@cs.cmu.edu, yuncheng@cs.cmu.edu}
\affiliation{
  \institution{Carnegie Mellon University}
  \city{Pittsburgh}
  \state{PA}
  \country{USA}
}

\title[{Multimodal Fusion Interactions: A Study of Human and Automatic Quantification}]{Multimodal Fusion Interactions:\\A Study of Human and Automatic Quantification}

\begin{abstract}
In order to perform multimodal fusion of heterogeneous signals, we need to understand their interactions: how each modality individually provides information useful for a task and how this information changes in the presence of other modalities. In this paper, we perform a comparative study of how humans annotate two categorizations of multimodal interactions: (1) \textit{partial labels}, where different annotators annotate the label given the first, second, and both modalities, and (2) \textit{counterfactual labels}, where the same annotator annotates the label given the first modality before asking them to explicitly reason about how their answer changes when given the second. We further propose an alternative taxonomy based on (3) \textit{information decomposition}, where annotators annotate the degrees of \textit{redundancy}: the extent to which modalities individually and together give the same predictions, \textit{uniqueness}: the extent to which one modality enables a prediction that the other does not, and \textit{synergy}: the extent to which both modalities enable one to make a prediction that one would not otherwise make using individual modalities. Through experiments and annotations, we highlight several opportunities and limitations of each approach and propose a method to automatically convert annotations of partial and counterfactual labels to information decomposition, yielding an accurate and efficient method for quantifying multimodal interactions.
\end{abstract}

\ccsdesc[500]{Computing methodologies~Machine Learning}

\keywords{Multimodal interactions; Multimodal fusion; Affective computing}

\maketitle

\begin{figure}[h]
\centering
\vspace{2mm}
\includegraphics[width=\linewidth]{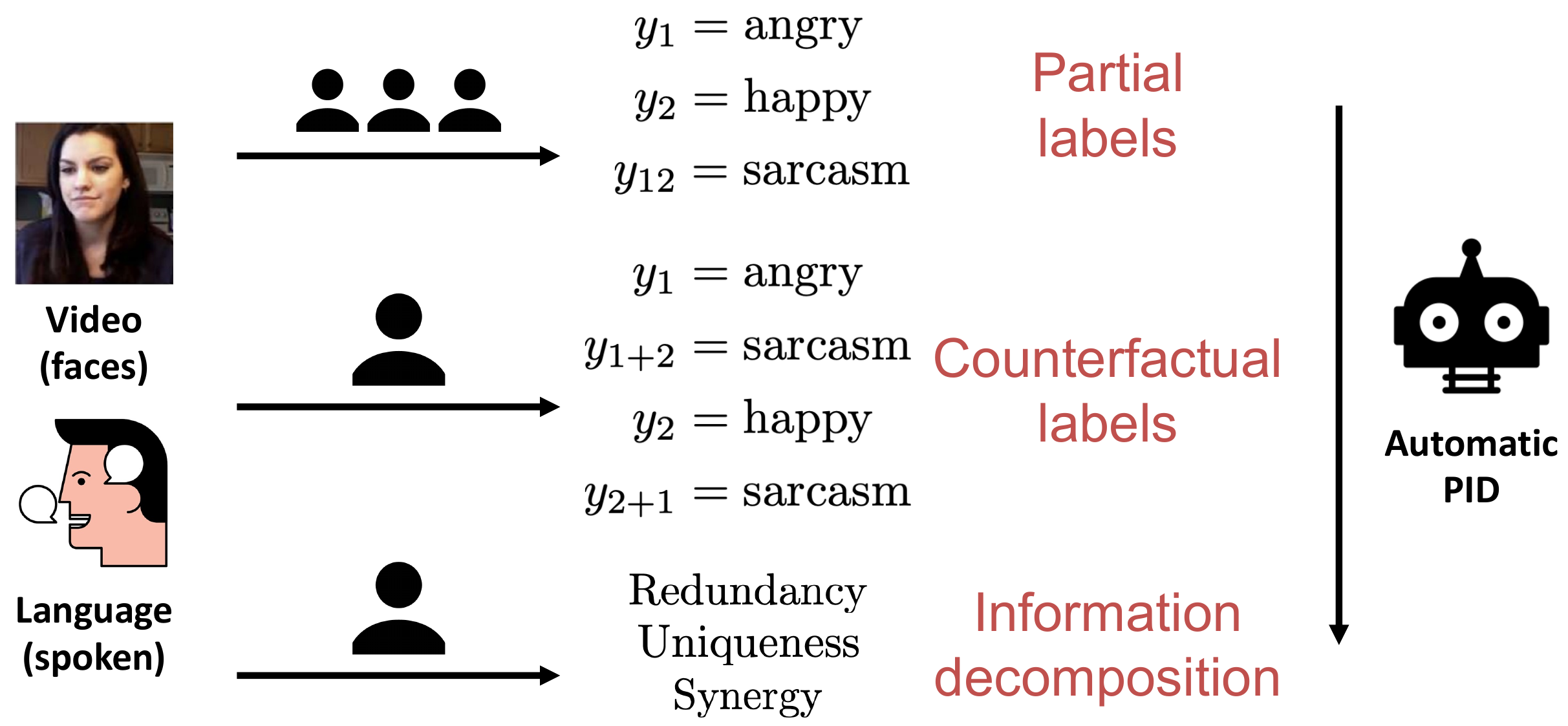}
\vspace{-2mm}
\caption{We study human annotation of multimodal fusion interactions via three categorizations: (1) \textit{partial labels} in which different randomly assigned annotators annotate the task given the first ($y_1$), second ($y_2$), and both modalities ($y_{12}$), (2) \textit{counterfactual labels}, where the same annotator is tasked to annotate the label given the first modality ($y_1$), before asking them to reason about how their answer chances when given the second ($y_{1+2}$), and vice versa ($y_2$ and $y_{2+1}$), and (3) \textit{information decomposition} where annotators annotate the degrees of modality redundancy, uniqueness, and synergy in predicting the task. Finally, we also propose a scheme based on PID~\cite{williams2010nonnegative} to automatically convert annotations of partial and counterfactual labels to information decomposition.}
\label{fig:overview}
\Description[Overview of human annotation schemes]{We study human annotation of multimodal fusion interactions via three categorizations: (1) partial labels in which different randomly assigned annotators annotate the task given the first, second, and both modalities, (2) counterfactual labels, where the same annotator is tasked to annotate the label given the first modality, before asking them to reason about how their answer chances when given the second, and vice versa, and (3) information decomposition where annotators annotate the degrees of modality redundancy, uniqueness, and synergy in predicting the task.}
\vspace{-4mm}
\end{figure}

\section{Introduction}

A core challenge in multimodal machine learning lies in understanding the ways that different modalities interact with each other in combination for a given prediction task~\cite{liang2022foundations}. We define the study of \textit{multimodal fusion interactions} as the categorization and measurement of how each modality individually provides information useful for a task and how this information changes in the presence of other modalities~\cite{provost2015umeme,d2015review,pantic2005affective}. Learning complex interactions are often quoted as motivation for many successful multimodal modeling paradigms in the machine learning and multimodal interaction communities, such as contrastive learning~\cite{radford2021learning,kim2021vilt}, modality-specific representations~\cite{tsai2019learning,yuan2021multimodal}, and higher-order interactions~\cite{zadeh2017tensor,liang2018multimodal,Jayakumar2020Multiplicative}.
Despite progress in new models that seem to better capture various interactions from increasingly complex real-world multimodal datasets~\cite{zadeh2017tensor,Jayakumar2020Multiplicative}, formally quantifying and measuring the interactions that are necessary to solve a multimodal task remains a fundamental research question~\cite{liang2022foundations,liang2023multiviz,hessel2020emap}.

In this paper, we perform a comparative study of how reliably human annotators can be leveraged to quantify different interactions in real-world multimodal datasets (see Figure~\ref{fig:overview}). We first start with a conventional method which we term \textit{partial labels}, where different randomly assigned annotators annotate the task given only the first modality ($y_1$), only the second modality ($y_2$), and both modalities ($y_{12}$)~\cite{provost2015umeme,d2015review,pantic2005affective,ruiz2006examining}. Beyond partial labels, we extend this idea to \textit{counterfactual labels}, where the same annotator is tasked to annotate the label given the first modality ($y_1$), before giving them the second modality and asking them to explicitly reason about how their answer changes ($y_{1+2}$), and vice versa ($y_2$ and $y_{2+1}$)~\cite{soleymani2011multimodal}.
Additionally, we propose an alternative taxonomy of multimodal interactions grounded in information theory~\cite{williams2010nonnegative,liang2023quantifying}, which we call \textit{information decomposition}: decomposing the total information two modalities provide about a task into \textit{redundancy}, the extent to which individual modalities and both in combination all give similar predictions on the task, \textit{uniqueness}, the extent to which the prediction depends only on one of the modalities and not the other, and \textit{synergy}, the extent to which task prediction changes with both modalities as compared to using either modality individually~\cite{williams2010nonnegative,liang2023quantifying}. Information decomposition has an established history in understanding feature interactions in neuroscience~\cite{timme2016high,wibral2017partial,pica2017quantifying,wibral2015bits}, physics~\cite{flecker2011partial,harder2013bivariate}, and biology~\cite{colenbier2020disambiguating,chan2017gene} since it exhibits desirable properties such as disentangling redundancy and synergy, normalization with respect to the total information two features provide towards a task, and established methods for automatic computation.

However, it remains a challenge to scale information decomposition to real-world high-dimensional and continuous modalities~\cite{liang2023quantifying,liang2023multimodal,bertschinger2014quantifying}, which has hindered its application in machine learning and multimodal interaction where complex video, audio, text, and other sensory modalities are prevalent. To quantify information decomposition for real-world multimodal tasks, we propose a new human annotation scheme where annotators provide estimates of redundancy, uniqueness, and synergy when presented with both modalities and the label. We find that this method works surprisingly well with strong annotator agreement and self-reported annotator confidence. Finally, given the promises of information decomposition~\cite{liang2023quantifying,Jayakumar2020Multiplicative,d2018affective}, we additionally propose a scheme to automatically convert annotations of partial and counterfactual labels to information decomposition using an information-theoretic method~\cite{williams2010nonnegative,bertschinger2014quantifying}, which makes it compatible with existing methods of annotating interactions~\cite{provost2015umeme,d2015review,pantic2005affective,ruiz2006examining}. Through comprehensive experiments on multimodal analysis of sentiment, humor, sarcasm, and question-answering, we compare these methods of quantifying multimodal interactions and summarize our key findings. We release our data and code at \url{https://github.com/pliang279/PID}.

\section{Related Work}

Multimodal fusion interactions have been studied based on the dimensions of response, information, and mechanics~\cite{liang2022foundations}. We define and highlight representative works in each category:

\begin{figure}[tbp]
\centering
\vspace{-0mm}
\includegraphics[width=0.9\linewidth]{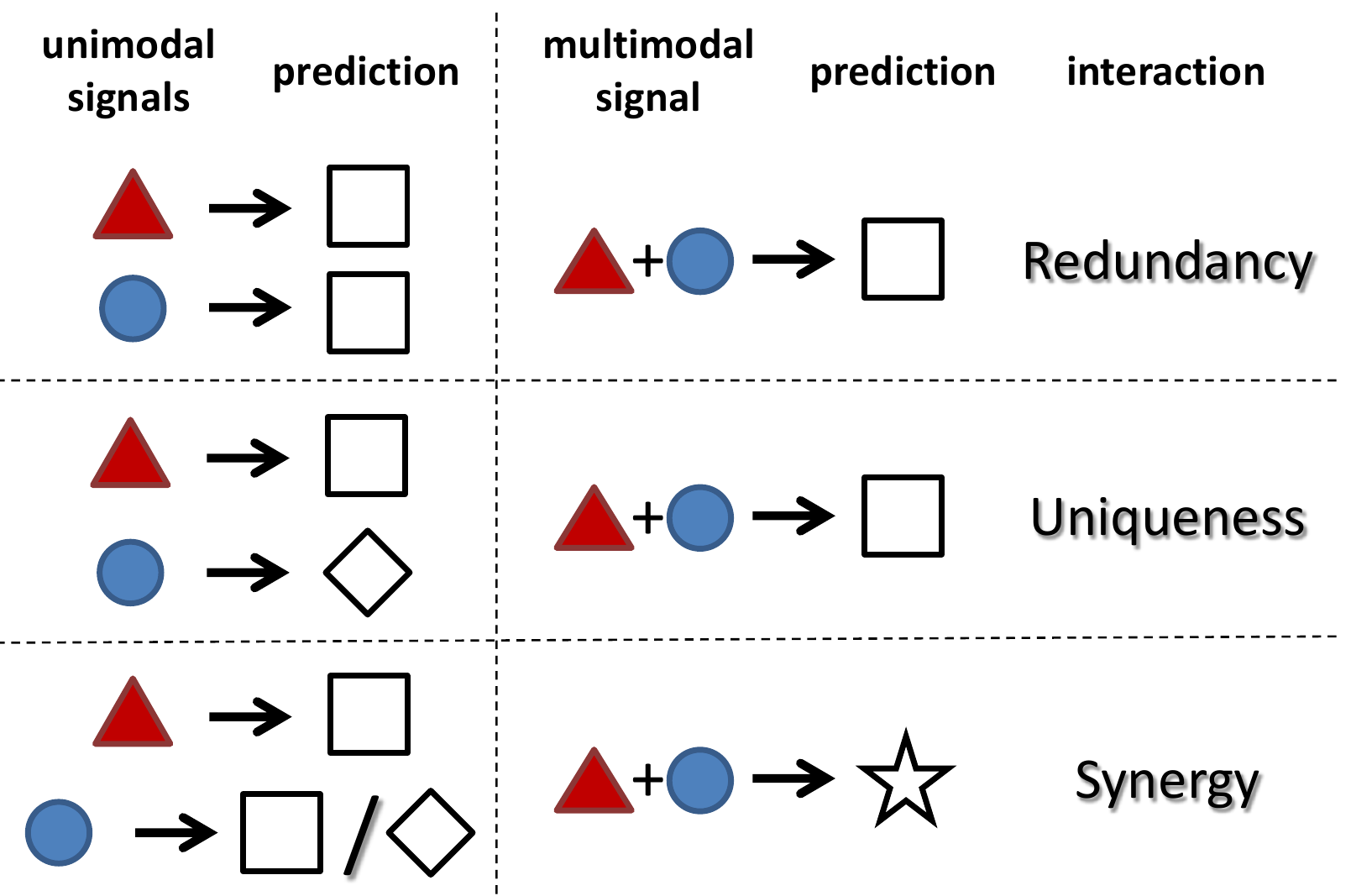}
\vspace{-2mm}
\caption{Categories of interaction response: redundancy happens when using either and both modalities give similar task predictions, uniqueness studies whether prediction depends on one of the modalities and not the other, and synergy measures how prediction changes with both modalities as compared to using either modality individually.}
\label{fig:definitions}
\Description[Definitions of multimodal interactions]{Categories of interaction response: redundancy happens when using either and both modalities give similar task predictions, uniqueness studies whether prediction depends on one of the modalities and not the other, and synergy measures how prediction changes with both modalities as compared to using either modality individually.}
\label{fig:definitions.}
\vspace{-4mm}
\end{figure}

\textbf{Interaction response} studies how the inferred response changes when two or more modalities are fused~\cite{liang2022foundations} (see Figure~\ref{fig:definitions}). For example, two modalities create a redundant response if the fused response is the same as responses from either modality or enhanced if the fused response displays higher confidence. Non-redundant interactions such as modulation or emergence can also happen~\citep{partan1999communication}. Many of these terms actually started from research in human and animal communicative modalities~\cite{partan1999communication,partan2005issues,flom2007development,ruiz2006examining} and multimedia~\cite{bateman2014text,marsh2003taxonomy}. Inspired by these ideas, a common measure of interaction response redundancy is defined as the distance between prediction logits using either feature~\cite{mazzetto21a}. This definition is also commonly used in minimum-redundancy feature selection~\cite{yu2003efficiently,yu2004efficient,auffarth2010comparison}. Research in multimedia has also categorized interactions into divergent, parallel, and additive~\cite{bateman2014text,kruk2019integrating,zhang2018equal}. Finally, human annotations have been leveraged to identify redundant modalities via a proxy of cognitive load~\cite{ruiz2006examining}. This paper primarily focuses on interaction response since it is the one easiest understood and annotated by humans, but coming up with formal definitions and measures of other interactions are critical directions for future work.

\textbf{Interaction information} investigates the nature of information overlap between multiple modalities. The information important for a task can be shared in both modalities, unique to one modality, or emerge only when both are present~\citep{liang2022foundations}. Information-theoretic measures naturally provide a mathematical formalism in the study of interaction information, for example through the mutual information between two variables~\cite{tosh2021contrastive,tian2020makes}. In the presence of two modalities and a label, extensions of mutual information to three variables, such as through total correlation~\cite{watanabe1960information,garner1962uncertainty} interaction information~\cite{mcgill1954multivariate,te1980multiple}, or partial information decomposition~\cite{williams2010nonnegative,bertschinger2014quantifying} have been proposed, and recent work has explored their estimation on large-scale real-world multimodal datasets~\cite{liang2023multimodal,liang2023quantifying}. From a semantic perspective, research in multimedia has studied various relationships that can exist between images and text~\cite{otto2020characterization,marsh2003taxonomy}, which has also inspired work in representing shared information through contrastive learning~\cite{radford2021learning}. While interaction information and response are naturally related, interaction response can be more fine-grained with respect to individual datapoints.

\begin{figure*}[tbp]
\centering
\begin{subfigure}{0.32\textwidth}
  \centering
  \includegraphics[width=\linewidth]{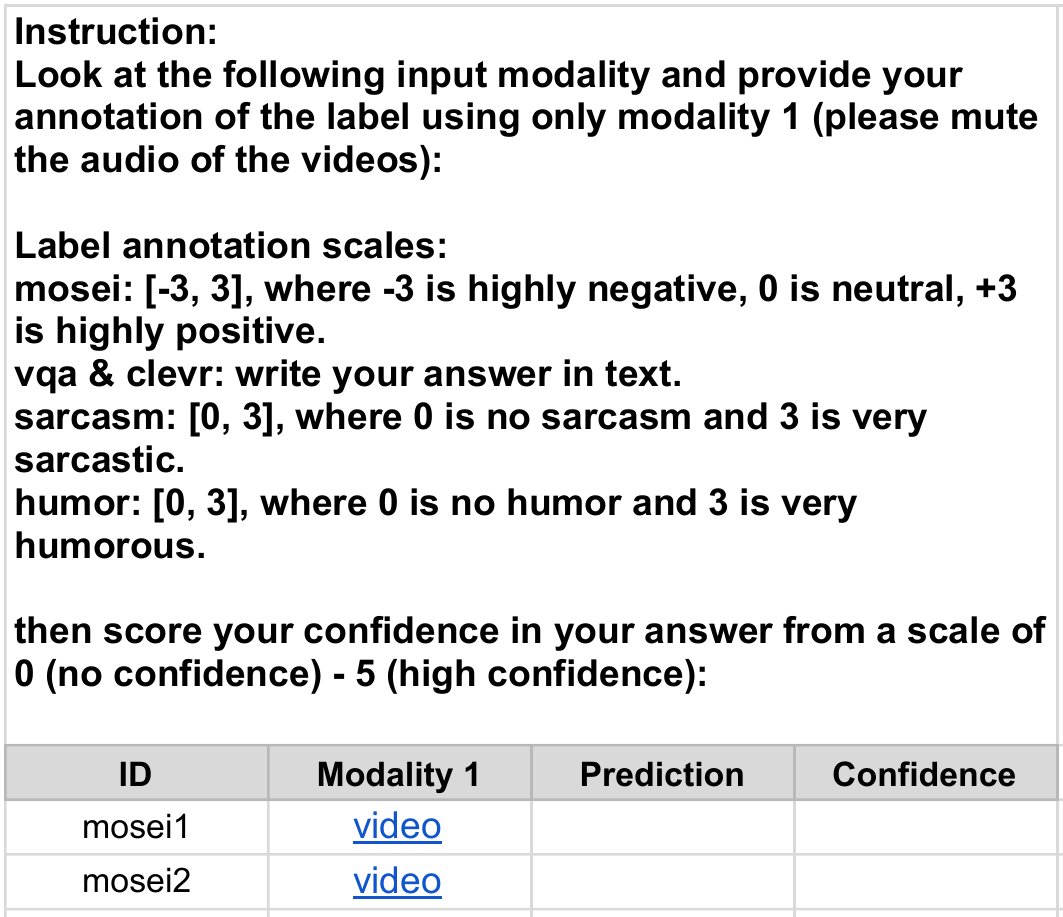}
  \caption{Sample user interface for annotating partial labels of the video modality.}
\end{subfigure}\hspace{4mm}
\begin{subfigure}{0.57\textwidth}
  \centering
  \includegraphics[width=\linewidth]{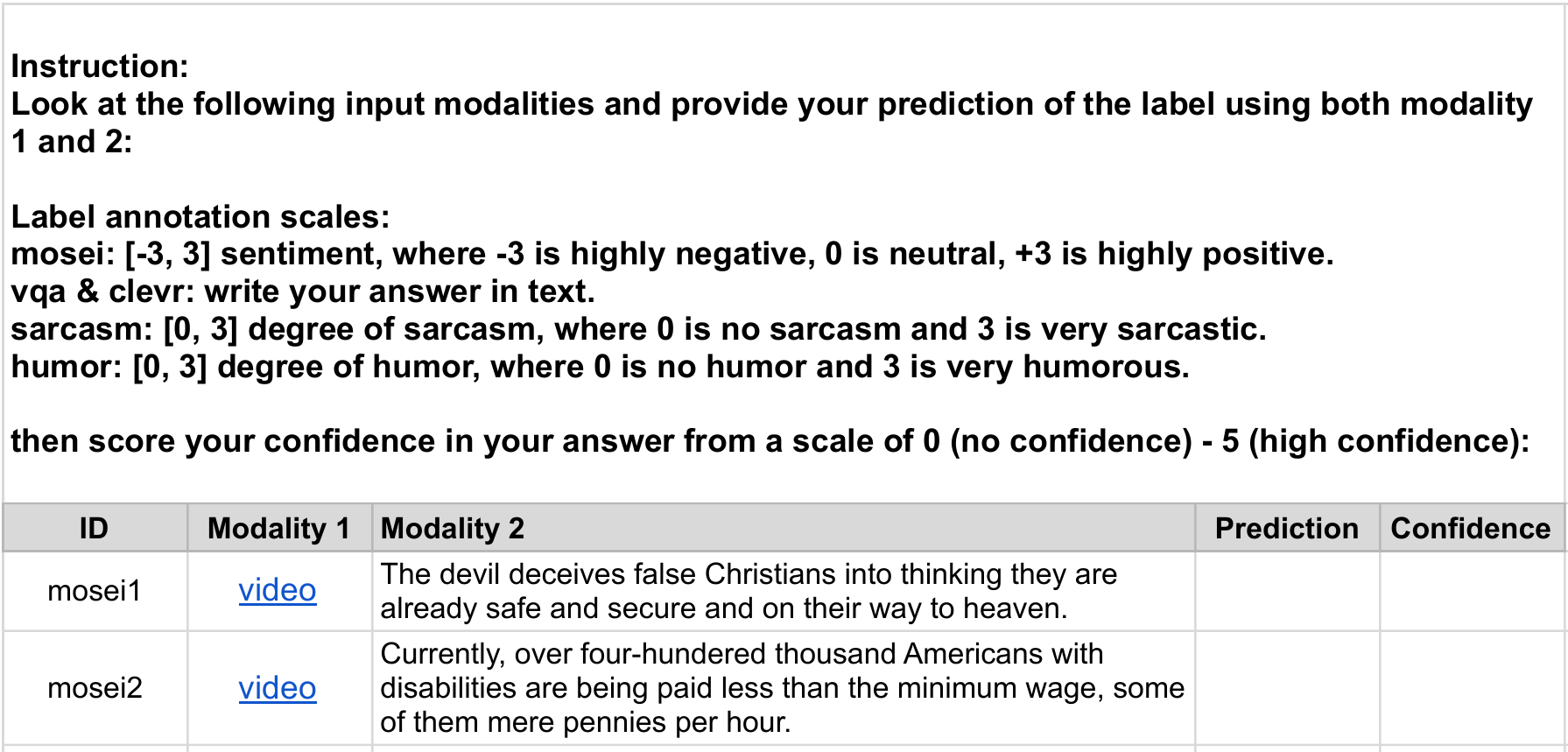}
  \caption{Sample user interface for annotating partial labels of the language modality.}
\end{subfigure}\vspace{2mm}\\%
\begin{subfigure}{\textwidth}
  \centering
  \includegraphics[width=\linewidth]{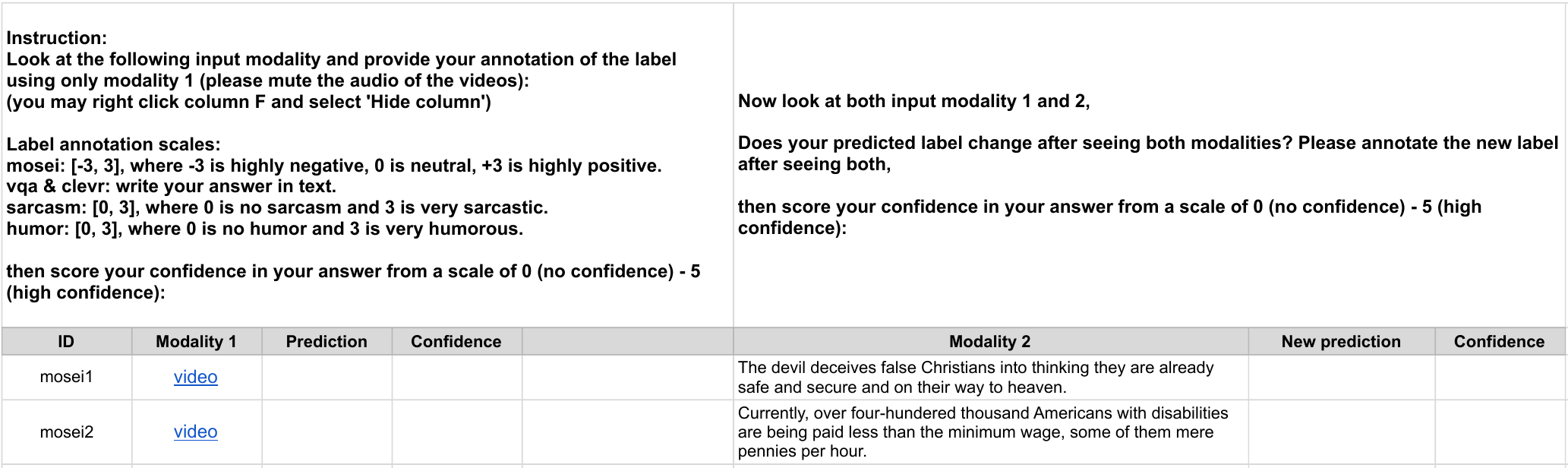}
  \caption{Sample user interface for annotating how the label changes from observing the video modality and then language by the same annotator.}
\end{subfigure}\vspace{2mm}\\%
\begin{subfigure}{\textwidth}
  \centering
  \includegraphics[width=\linewidth]{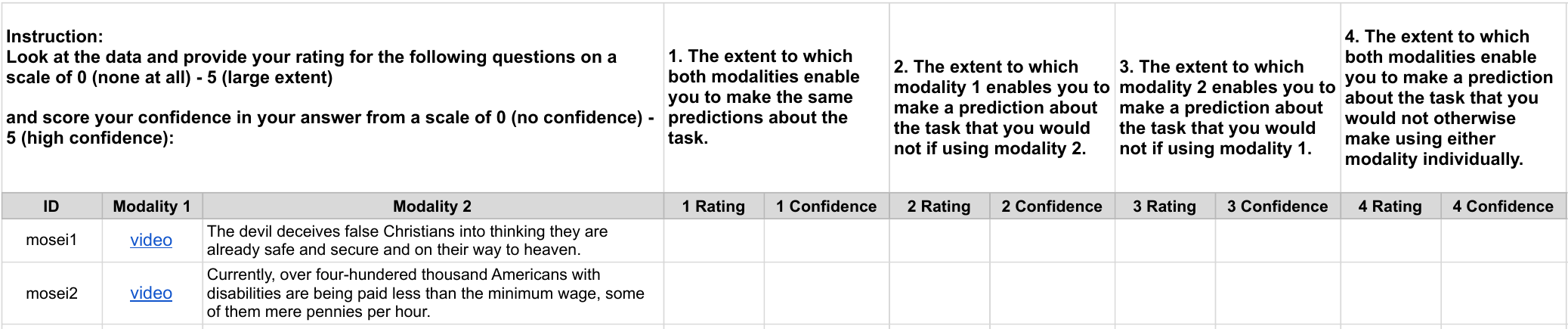}
  \caption{Sample user interface for annotating information decomposition of redundancy, uniqueness, and synergy.}
\end{subfigure}\vspace{2mm}\\%
\vspace{-4mm}
\caption{User interfaces for annotating partial labels (a, b), counterfactual labels (c), and information decomposition (d).}
\label{fig:annotations}
\Description[Screenshots of user annotation interfaces]{We show sample user interfaces for annotating partial labels for video and language modalities, counterfactual labels for how the label changes from observing the video modality and then language, and for annotating information decomposition of redundancy, uniqueness, and synergy.}
\vspace{-2mm}
\end{figure*}

Finally, the study of \textbf{interaction mechanics} studies how mathematical operators can be used to capture interactions during multimodal fusion. For example, interaction mechanics can be expressed as additive~\cite{friedman2008predictive}, multiplicative~\cite{Jayakumar2020Multiplicative}, tensor~\cite{zadeh2017tensor}, non-linear~\cite{ngiam2011multimodal}, and recurrent~\cite{liang2018multimodal} forms, as well as logical, causal, or temporal operations~\cite{unsworth2014multimodality}. By making assumptions on a specific functional form of interactions (e.g., additive vs non-additive), prior work has been able to quantify their presence or absence~\cite{sorokina2008detecting,tsang2018detecting,tsang2019feature} in real-world multimodal datasets and models through studies of architecture-specific attention and parameter weights~\cite{}, model-agnostic gradient-based visualizations~\cite{lyu2022dime,wang2021m2lens,liang2023multiviz}, and projections into simpler models~\cite{hessel2020emap,wortwein2022beyond}.

\vspace{-1mm}
\section{\mbox{Annotating Multimodal Interactions}}

In order to study interaction response during multimodal fusion, we first review the estimation of partial labels via random assignment, before discussing an alternative approach through counterfactual labels. Finally, we motivate information decomposition into redundancy, uniqueness, and synergy, which offers a different perspective and new benefits for studying multimodal interactions.

\vspace{-1mm}
\subsection{\mbox{Annotating partial labels}}

The standard approach involves tasking randomly assigned annotators to label their prediction of the label when presented with only the first modality ($y_1$), the label when presented with only the second modality ($y_2$), and the label when presented with both modalities ($y_{12}$)~\cite{provost2015umeme,d2015review,pantic2005affective}. Annotators are typically randomly assigned to each modality so that their labeling process is not influenced by observing other modalities, resulting in independently annotated partial labels. In this setup, the instructions given are:
\begin{enumerate}[noitemsep,topsep=0pt,nosep,leftmargin=*,parsep=0pt,partopsep=0pt]
    \item $y_1$: Show modality 1, and ask the annotator to predict the label.
    \item $y_2$: To another annotator, show only modality 2, and ask the annotator to predict the label.
    \item $y_{12}$: To yet another annotator, show both modalities, and ask the annotator to predict the label.
\end{enumerate}
After reporting each partial label, the annotators are also asked to report confidence on a 0-5 scale (0: no confidence, 5: high confidence). We show a screenshot of a sample user interface in Figure~\ref{fig:annotations} (top) and provide more annotation details in Appendix~\ref{app:indirect1}.

\subsection{\mbox{Annotating counterfactual labels}}

As another alternative to random assignment, we draw insight from counterfactual estimation where the same annotator annotates the label given a single modality, before giving them the second modality and asking them to reason about how their answer changes. The instructions provided to the first annotator are:
\begin{enumerate}[noitemsep,topsep=0pt,nosep,leftmargin=*,parsep=0pt,partopsep=0pt]
    \item $y_1$: Show modality 1, and ask them to predict the label.
    \item $y_{1+2}$: Now show both modalities and ask if their predicted label explicitly changes after seeing both modalities.
\end{enumerate}
To a separate annotator, we provide the following instructions:
\begin{enumerate}[noitemsep,topsep=0pt,nosep,leftmargin=*,parsep=0pt,partopsep=0pt]
    \item $y_2$: Show modality 2, and ask them to predict the label.
    \item $y_{2+1}$: Now show both modalities and ask if their predicted label explicitly changes after seeing both modalities.
\end{enumerate}
The annotators also report confidence on a 0-5 scale (see sample user interface in Figure~\ref{fig:annotations} (middle) and exact annotation procedures in Appendix~\ref{app:indirect2}). While the first method by random assignment estimates the average effect of each modality on the label as is commonly done in randomized control trials~\cite{benson2000comparison} (since estimates of partial labels for each modality are done separately in expectation over all users), this counterfactual approach measures the actual causal effect of seeing the second modality towards the label for the same user~\cite{kaushik2019learning,abbasnejad2020counterfactual,goyal2019counterfactual}.

\subsection{\mbox{Annotating information decomposition}}
\label{sec:direct_human}

Finally, we propose an alternative categorization of multimodal interactions based on information theory, which we call \textit{information decomposition}: decomposing the total information two modalities provide about a task into \textit{redundancy}, the extent to which individual modalities and both in combination all give similar predictions on the task, \textit{uniqueness}, the extent to which the prediction depends only on one of the modalities and not the other, or \textit{synergy}, the extent to which task prediction changes with both modalities as compared to using either modality individually~\cite{williams2010nonnegative,liang2023quantifying}.

This view of interactions is useful since it has a formal grounding in information theory~\cite{shannon1948mathematical} and information decomposition~\cite{williams2010nonnegative}. Information theory formalizes the amount of information that a variable ($X$) provides about another ($Y$), and is quantified by Shannon's mutual information (MI):
\begin{align}
    I(X_1; X_2) = \int p(x_1,x_2) \log \frac{p(x_1,x_2)}{p(x_1) p(x_2)} d x_1 d x_2,
\end{align}
which measures the amount of information (in bits) obtained about $X_1$ by observing $X_2$. By extension, conditional MI is the expected value of the MI of two random variables (e.g., $X_1$ and $X_2$) given the value of a third (e.g., $Y$):
\begin{align}
    I(X_1;X_2|Y) = \int p(x_1,x_2,y) \log \frac{p(x_1,x_2|y)}{p(x_1|y) p(x_2|y)} d x_1 d x_2 d y.
\end{align}

\subsubsection{Multivariate information theory}

While information theory works well in two variables, the extension of information theory to measure redundancy and other interactions requires its extension to three or more variables, which remains an open challenge. The most natural extension, through interaction information~\cite{mcgill1954multivariate,te1980multiple}, has often been indirectly used as a measure of redundancy in co-training~\cite{blum1998combining,balcan2004co,christoudias2008multi} and multi-view learning~\cite{tosh2021contrastive,tsai2020self,tian2020makes,sridharan2008information}. It is defined for three variables as the difference in mutual information and conditional mutual information:
\begin{align}
    I(X_1; X_2; Y) = I(X_1; X_2) - I(X_1; X_2|Y),
\end{align}
and can be defined inductively for more than three variables. However, interaction information has some significant shortcomings: $I(X_1; X_2; Y)$ can be both positive and negative, leading to considerable difficulty in its interpretation when redundancy as an information quantity is negative~\cite{jakulin2003quantifying,liang2023quantifying}. Furthermore, the total information is only equal to redundancy and uniqueness ($I(X_1,X_2; Y) = I(X_1; X_2; Y) + I(X_1; Y|X_2) + I(X_2; Y|X_1)$), and there is no measurement of synergy in this framework.

\subsubsection{Information decomposition}

\begin{figure}[tbp]
\centering
\vspace{-0mm}
\includegraphics[width=0.7\linewidth]{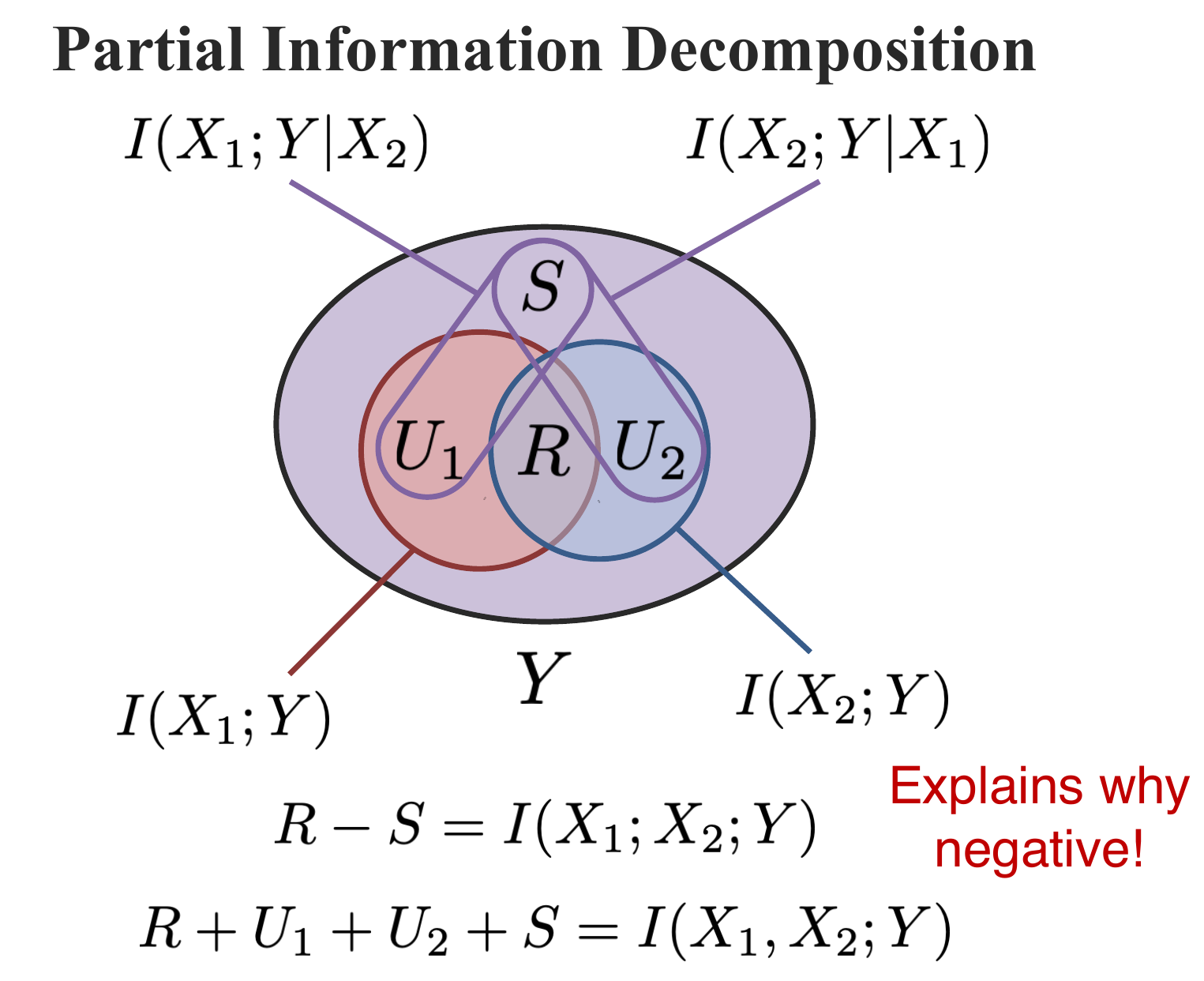}
\vspace{-4mm}
\caption{Partial information decomposition gives a principled way to estimate the interactions that are redundant between $2$ modalities, unique to one modality, and synergistic only when both modalities are present.}
\label{fig:info}
\Description[Illustration of partial information decomposition]{A venn diagram showing the areas depicting information redundant between 2 modalities, unique to one modality, and synergistic only when both modalities are present.}
\vspace{-4mm}
\end{figure}

Partial information decomposition (PID)~\cite{williams2010nonnegative} was designed to solve some of the issues with multivariate information theory. PID is a class of definitions for redundancy $R$ between $X_1$ and $X_2$, unique information $U_1$ in $X_1$ and $U_2$ in $X_2$, and synergy $S$ when both $X_1$ and $X_2$ are present such that the following equations hold (see Figure~\ref{fig:info} for a visual depiction):
\begin{alignat}{3}
\label{eqn:consistency}
    R + U_1 &= I(X_1; Y), &&R + U_2 &&= I(X_2; Y),\\
    U_1 + S &= I(X_1; Y | X_2), \quad &&U_2 + S &&= I(X_2; Y | X_1),\\
    R - S &= I(X_1; X_2; Y). \label{eqn:consistency2}
\end{alignat}
PID resolves the issue of negative $I(X_1; X_2; Y)$ in conventional information theory by separating $R$ and $S$ such that $R-S = I(X_1; X_2; Y)$, identifying that prior redundancy measures confound actual redundancy and synergy. Furthermore, if $I(X_1; X_2; Y) = 0$ then existing frameworks are unable to distinguish between positive values of true $R$ and $S$ canceling each other out, while PID separates and can estimate non-zero (but equal) values of both $R$ and $S$.

\subsubsection{Annotating information decomposition}

\begin{figure*}[tbp]
\centering
\vspace{-0mm}
\includegraphics[width=0.9\linewidth]{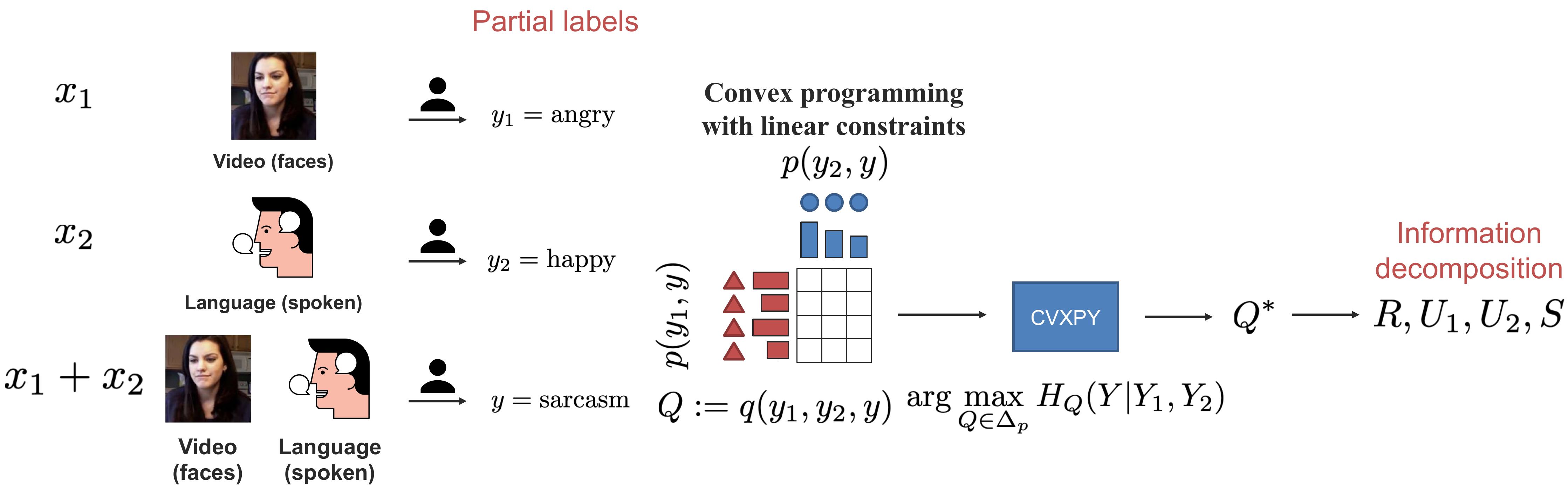}
\vspace{-1mm}
\caption{Overview of our proposed method to convert partial or counterfactual labels to information decomposition values. We treat the dataset of partial labels $\mathcal{D} = \{(y_1,y_2,y)_{i=1}^n\}$ as a joint distribution with $y_1$ and $y_2$ as `multimodal inputs' and the target label $y$ as the `output'. Estimating response redundancy, uniqueness, and synergy then boils down to solving a convex optimization problem with marginal constraints, which can be done accurately and efficiently. This method is applicable to many annotated multimodal datasets and yields consistent, comparable, and standardized interaction estimates.}
\label{fig:est1}
\Description[Overview of our proposed method to convert partial or counterfactual labels to information decomposition values]{We convert partial or counterfactual labels to information decomposition values via a convex optimization method with marginal constraints, which can be done accurately and efficiently. This method is applicable to many annotated multimodal datasets and yields consistent, comparable, and standardized interaction estimates.}
\vspace{-2mm}
\end{figure*}

While information decomposition has a formal definition and exhibits nice properties, it remains a challenge to scale information decomposition to real-world high-dimensional and continuous modalities~\cite{liang2023quantifying,bertschinger2014quantifying}. To quantify information decomposition for real-world tasks, we investigate whether human judgment can be used as a reliable estimator. We propose a new annotation scheme where we show both modalities and the label and ask each annotator to annotate the degree of redundancy, uniqueness, and synergy on a scale of 0-5 using the following definitions inspired by the formal definitions in information decomposition:
\begin{enumerate}[noitemsep,topsep=0pt,nosep,leftmargin=*,parsep=0pt,partopsep=0pt]
    \item $R$: The extent to which using the modalities individually and together gives the same predictions on the task,
    \item $U_1$: The extent to which $x_1$ enables you to make a prediction about the task that you would not if using $x_2$,
    \item $U_2$: The extent to which $x_2$ enables you to make a prediction about the task that you would not if using $x_1$,
    \item $S$: The extent to which only both modalities enable you to make a prediction about the task that you would not otherwise make using either modality individually,
\end{enumerate}
alongside their confidence in their answers on a scale of 0-5. We show a sample user interface for the annotations in Figure~\ref{fig:annotations} (bottom) and include exact annotation procedures in Appendix~\ref{app:direct}.

\vspace{-1mm}
\section{Converting Partial Labels to PID}

Finally, we propose a method to automatically convert partial labels, which are present in many existing multimodal datasets~\cite{provost2015umeme,d2015review,pantic2005affective}, into information decomposition interaction values. Define the multimodal label $y$ as $y_{12}$ in the case of partial labels and the average of $y_{1+2}$ and $y_{2+1}$ in the case of counterfactual labels. Then, the partial and counterfactual labels are related to redundancy, uniqueness, and synergy in the following ways:
\begin{enumerate}[noitemsep,topsep=0pt,nosep,leftmargin=*,parsep=0pt,partopsep=0pt]
    \item $R$ is high when $y_1$, $y_2$, and $y$ are all close to each other,
    \item $U_1$ is high when $y_1$ is close to $y$ but $y_2$ is far from $y$,
    \item $U_2$ is high when $y_2$ is close to $y$ but $y_1$ is far from $y$,
    \item $S$ is high when $y_1, y_2$ are both far from $y$.
\end{enumerate}

While these partial labels are intuitively related to information decomposition, coming up with a concrete equation to convert $y_1$, $y_2$, and $y$ to actual interaction values is surprisingly difficult and involves many design decisions. For example, what distance measure do we use to measure closeness in label space? Furthermore, computing $R$ depends on 3 distances, $U_1$ and $U_2$ depend on 2 distances but inversely on 1 distance, and $S$ depends on 2 distances. How do we obtain interaction values that lie on comparable scales so that they can be compared reliably?

\begin{table*}[]
\fontsize{9}{11}\selectfont
\setlength\tabcolsep{2.0pt}
\vspace{1mm}
\caption{Collection of datasets used for our study of multimodal fusion interactions covering diverse modalities, tasks, and research areas in multimedia and affective computing.}
\centering
\vspace{1mm}
\begin{tabular}{l|ccccc}
\hline \hline
Datasets & Modalities & Size & Prediction task & Research Area \\
\hline
\textsc{VQA 2.0}~\cite{goyal2017making} & $\{ \textrm{image}, \textrm{question} \}$ & $1,100,000$ & QA & Multimedia \\
\textsc{CLEVR}~\cite{johnson2017clevr} & $\{ \textrm{image}, \textrm{question} \}$ & $853,554$ & QA & Multimedia \\
\textsc{MOSEI}~\cite{zadeh2018multimodal} & $\{ \textrm{text}, \textrm{video}, \textrm{audio} \}$ & $22,777$ & sentiment, emotions & Affective Computing \\
\textsc{UR-FUNNY}~\cite{hasan2019ur} & $\{ \textrm{text}, \textrm{video}, \textrm{audio} \}$ & $16,514$ & humor & Affective Computing \\
\textsc{MUStARD}~\cite{castro2019towards} & $\{ \textrm{text}, \textrm{video}, \textrm{audio} \}$ & $690$ & sarcasm & Affective Computing \\
\hline \hline
\end{tabular}

\vspace{-2mm}
\label{tab:setup}
\end{table*}

\subsection{Automatic conversion}

Our key insight is that the aforementioned issues are exactly what inspired much of the research in information theory and decomposition in the first place: in information theory, the lack of a distance measure is solved by working with probability distributions where information-theoretic distances like KL-divergence are well-defined and standardized, the issues of normalization are solved using a standardized unit of measure (bits in log-base 2), and the issues of incomparable scales are solved by the consistency equations (\ref{eqn:consistency})-(\ref{eqn:consistency2}) relating PID values to each other and to the total task-relevant information in both modalities.

Armed with these formalisms of information theory and information decomposition, we propose a method to convert human-annotated partial predictions into redundancy, uniqueness, and synergy (see Figure~\ref{fig:est1} for an overview). To do so, we treat the dataset of partial predictions $\mathcal{D} = \{(y_1,y_2,y)_{i=1}^n\}$ as a joint distribution with $y_1$ and $y_2$ as `multimodal inputs' sampled over the label support $\mathcal{Y}$, and the target label $y$ as the `output' also over $\mathcal{Y}$. Following this, we adopt a precise definition of redundancy, uniqueness, and synergy used by~\citet{bertschinger2014quantifying}, where the interactions are defined as the solution to the optimization problems:
\begin{align}
    R &= \max_{q \in \Delta_p} I_q(Y_1; Y_2; Y), \label{eqn:R-def}\\
    U_1 &= \min_{q \in \Delta_p} I_q(Y_1; Y | Y_2), \quad U_2 = \min_{q \in \Delta_p} I_q(Y_2; Y| Y_1), \label{eqn:U2-def}\\
    S &= I_p(Y_1,Y_2; Y) - \min_{q \in \Delta_p} I_q(Y_1,Y_2; Y), \label{eqn:S-def}
\end{align}
where $\Delta_p = \{ q \in \Delta: q(y_i,y)=p(y_i,y) \ \forall y_i,y\in\mathcal{Y}, i \in \{1,2\} \}$ and the notation $I_p(\cdot)$ and $I_q(\cdot)$ disambiguates MI under joint distributions $p$ and $q$ respectively. The key difference in this definition of PID lies in optimizing $q \in \Delta_p$ to satisfy the marginals $q(y_i,y)=p(y_i,y)$, but relaxing the coupling between $y_1$ and $y_2$: $q(y_1,y_2)$ need not be equal to $p(y_1,y_2)$. The intuition behind this is that one should be able to infer redundancy and uniqueness given only access to separate marginals $p(y_1,y)$ and $p(y_2,y)$, and therefore they should only depend on $q \in \Delta_p$ which match these marginals. Synergy, however, requires knowing the coupling $p(y_1,y_2)$, and this is reflected in equation (\ref{eqn:S-def}) depending on the full $p$ distribution.

\subsection{\mbox{Estimating information decomposition}}

These optimization problems can be solved accurately and efficiently using convex programming. Importantly, the $q^*$ that solves (\ref{eqn:R-def})-(\ref{eqn:S-def}) can be rewritten as the solution to the max-entropy optimization problem: $q^* = \argmax_{q \in \Delta_p} H_q(Y | Y_1, Y_2)$~\cite{bertschinger2014quantifying}. Since the support of the label space $\mathcal{Y}$ is usually small and discrete for classification, or small and continuous for regression, we can represent all valid joint distributions $q(y_1,y_2,y)$ as a set of tensors $Q$ of shape $|\mathcal{Y}| \times |\mathcal{Y}| \times |\mathcal{Y}|$ with each entry representing $Q[i,j,k] = q(Y_1=i,Y_2=j,Y=k)$. The problem then boils down to optimizing over tensors $Q$ that are valid joint distributions and that match marginals over each modality and the label (i.e., making sure $Q \in \Delta_p$).

Given a tensor parameter $Q$, our objective is $H_Q(Y | Y_1, Y_2)$, which is concave. This is therefore a convex optimization problem and the marginal constraints can be written as linear constraints. Given a dataset $D = \{(y_1,y_2,y)_{i=1}^n\}$, $p(y_1,y)$ and $p(y_2,y)$ are first estimated before enforcing $q(y_1,y) = p(y_1,y)$ and $q(y_2,y) = p(y_2,y)$ through linear constraints: the 3D-tensor $Q$ summed over the second dimension gives $q(y_1,y)$ and summed over the first dimension gives $q(y_2,y)$. Our final optimization problem is given by
\begin{align}
    &\argmax_{Q} H_Q(Y | Y_1, Y_2), \\
    \textrm{such that} \quad &\sum_{y_2} Q = p(y_1,y) \quad \sum_{y_1} Q = p(y_2,y), \\
    &Q \ge 0, \sum_{y_1,y_2,y} Q = 1.
    \label{eqn:cvx-optimizer}
\end{align}
Since this is a convex optimization problem with linear constraints, CVXPY~\cite{diamond2016cvxpy} returns the exact answer $Q^*$ efficiently. Plugging the learned $Q^*$ into equations (\ref{eqn:R-def})-(\ref{eqn:S-def}) yields the desired estimates for redundancy, uniqueness, and synergy.

Therefore, this estimator can automatically convert partial or counterfactual labels annotated by humans in existing multimodal datasets~\cite{provost2015umeme,d2015review,pantic2005affective} into information decomposition interactions, yielding consistent, comparable, and standardized estimates.

\section{Experiments}

In this section, we design experiments to compare the annotation of multimodal interactions via randomized partial labels, counterfactual labels, and information decomposition into redundancy, uniqueness, and synergy.

\subsection{Experimental setup}

\subsubsection{Datasets and tasks}

Our experiments involve a large collection of datasets spanning the language, visual, and audio modalities across affective computing and multimedia. We summarize the datasets used in Table~\ref{tab:setup} and provide more details here:

\textbf{1. \textsc{VQA 2.0}}~\cite{goyal2017making} is a balanced version of the popular VQA~\cite{agrawal2017vqa} dataset by collecting complementary images such that every question is associated with a pair of similar images that result in two different answers to the same question. This reduces the occurrence of spurious correlations in the dataset and enables the training of more robust models.

\textbf{2. \textsc{CLEVR}}~\cite{johnson2017clevr} is a dataset for studying the ability of multimodal systems to perform visual reasoning. It contains $100,000$ rendered images and about $853,000$ unique automatically generated questions that test visual reasoning abilities such as counting, comparing, logical reasoning, and memory.

\textbf{3. \textsc{MOSEI}}~\cite{zadeh2018multimodal} is a collection of $22,000$ opinion video clips annotated with labels for subjectivity and sentiment intensity. The dataset includes per-frame, and per-opinion annotated visual features, and per-milliseconds annotated audio features. Sentiment intensity is annotated in the range $[-3,+3]$. Videos are collected from YouTube with a focus on video blogs which reflect real-world speakers expressing their behaviors through monologue videos.

\textbf{4. \textsc{UR-FUNNY}}~\citep{hasan2019ur}: Humor is an inherently multimodal communicative tool involving the effective use of words (text), accompanying gestures (visual), and prosodic cues (acoustic). UR-FUNNY consists of more than $16,000$ video samples from TED talks annotated for humor, and covers speakers from various backgrounds, ethnic groups, and cultures.

\textbf{5. \textsc{MUStARD}}~\citep{castro2019towards} is a multimodal video corpus for research in sarcasm detection compiled from popular TV shows including Friends, The Golden Girls, The Big Bang Theory, and Sarcasmaholics Anonymous. \textsc{MUStARD} consists of $690$ audiovisual utterances annotated with sarcasm labels. Sarcasm requires careful modeling of complementary information, particularly when the information from each modality does not agree with each other.

Overall, the datasets involved in our experiments cover diverse modalities such as images, video, audio, and text, with prediction tasks spanning humor, sarcasm sentiment, emotions, and question-answering from affective computing and multimedia.

\subsubsection{Annotation details}

Participation in all annotations was fully voluntary and we obtained consent from all participants prior to annotations. The authors manually took anonymous notes on all results and feedback in such a manner that the identities of annotators cannot readily be ascertained directly or through identifiers linked to the subjects. 
Participants were not the authors nor in the same research groups as the authors, but they all hold or are working towards a graduate degree in a STEM field and have knowledge of machine learning. None of the participants knew about this project before their session and each participant only interacted with the setting they were involved in.

We sample $50$ datapoints from each of the $5$ datasets in Table~\ref{tab:setup} and give them to a total of $12$ different annotators:
\begin{itemize}[noitemsep,topsep=0pt,nosep,leftmargin=*,parsep=0pt,partopsep=0pt]
	\item 3 annotators for direct annotation of interactions,
	\item 3 annotators for partial labeling of $y_1$, $y_2$, and $y_{12}$,
	\item 3 annotators for counterfactual, labeling $y_1$ first then $y_{1+2}$,
	\item 3 annotators for counterfactual, labeling $y_2$ first then $y_{2+1}$.
\end{itemize}

We summarize the results and key findings:

\subsection{\mbox{Annotating partial and counterfactual labels}}

\begin{table}[t]
\centering
\fontsize{9}{11}\selectfont
\setlength\tabcolsep{3.0pt}
\vspace{1mm}
\caption{Annotator agreement and average confidence scores of human annotations for: (1) partial labels via random assignment, (2) counterfactual labels, and (3) information decomposition into redundancy, uniqueness, and synergy.}
\centering

\begin{tabular}{l|ccc|ccc}
\hline \hline
Task & \multicolumn{3}{c|}{Agreement} & \multicolumn{3}{c}{Confidence} \\
\hline
Measure & $y_1$ & $y_2$ & $y_{12}$ & $y_1$ & $y_2$ & $y_{12}$ \\
\hline
Partial labels & $0.53$ & $0.73$ & $0.72$ & $2.96$ & $2.17$ & $4.68$ \\
\hline \hline
\end{tabular}

\vspace{1mm}

\begin{tabular}{l|cccc|cccc}
\hline \hline
Task & \multicolumn{4}{c|}{Agreement} & \multicolumn{4}{c}{Confidence} \\
\hline
Measure & $y_1$ & $y_{1+2}$ & $y_2$ & $y_{2+1}$ & $y_1$ & $y_{1+2}$ & $y_2$ & $y_{2+1}$ \\
\hline
Counterfactual & $0.44$ & $0.69$ & $0.52$ & $0.70$ & $2.19$ & $4.31$ & $2.33$ & $4.53$ \\
\hline \hline
\end{tabular}

\vspace{1mm}

\begin{tabular}{l|cccc|cccc}
\hline \hline
Task & \multicolumn{4}{c|}{Agreement} & \multicolumn{4}{c}{Confidence} \\
\hline
Measure & \red & \uone & \utwo & \syn & \red & \uone & \utwo & \syn \\
\hline
Info. decomposition & $0.43$ & $0.47$ & $0.54$ & $0.49$ & $4.51$ & $4.38$ & $4.46$ & $4.48$ \\
\hline \hline
\end{tabular}

\vspace{-0mm}
\label{tab:agreement}
\end{table}

We show the agreement scores of partial and counterfactual labels in Table~\ref{tab:agreement} and note some observations below:
\begin{itemize}[noitemsep,topsep=0pt,nosep,leftmargin=*,parsep=0pt,partopsep=0pt]

    \item \textbf{Comparing partial with counterfactual labels}: Counterfactual label agreement ($0.70$) is similar to randomized label agreement ($0.72$). In particular, annotating the video-only modality ($y_1$) for video datasets in the randomized setting appears to be confusing with an agreement of only $0.51$. We hypothesize that this is due to the challenge of detecting sentiment, sarcasm, and humor in videos without audio and when no obvious facial expression or body language is shown. Furthermore, we observe similar confidence in predicting the label when adding the second modality in the counterfactual setting versus showing both modalities upfront in the randomized setting: $4.42$ vs $4.68$.
    
    \item \textbf{Agreement and confidence datasets}: We examined the agreement for each dataset in the randomized and counterfactual settings respectively. In both settings, we found \textsc{MOSEI} is the easiest dataset with the highest agreement of $0.75$, $0.60$, $0.65$ for annotating $y_1$, $y_2$, and $y_{12}$ and $0.88$, $0.66$, $0.83$, $0.91$ for annotating $y_1$, $y_{1+2}$, $y_2$, and $y_{2+1}$. Meanwhile, \textsc{MUStARD} is the hardest, with the agreement as low as $-0.21$, $0.04$, and $0.17$ in the randomized setting. The average confidence for annotating partial labels is actually high (above $3.5$) for all datasets except unimodal predictions for \textsc{VQA} and \textsc{CLEVR}, which is as low as $0.43$ and $0.33$. This is understandable since these two image-based question-answering tasks are quite synergistic and cannot be performed using only one of the modalities, whereas annotator confidence when seeing both modalities is a perfect $5/5$.

    \item \textbf{Effect of counterfactual order}: Apart from a slight decrease in agreement in labeling $y_1$ first then $y_{1+2}$ and the slight increase in agreement in $y_2$ then $y_{2+1}$, we do not observe a significant difference caused by the counterfactual order. This is confirmed by the qualitative feedback from annotators: one responded that they found no difference between both orders and gave mostly similar responses to both.
\end{itemize}
Overall, we find that while both partial and counterfactual labels are reasonable choices for quantifying multimodal interactions, the annotation of counterfactual labels yields higher agreement and confidence than partial labels via random assignment.

\vspace{-1mm}
\subsection{\mbox{Annotating information decomposition}}

\begin{table*}[t]
\centering
\fontsize{9}{11}\selectfont
\setlength\tabcolsep{2.5pt}
\vspace{1mm}
\caption{Comparing (1) direct human annotation of information decomposition, (2) converting human-annotated partial labels to interactions via PID, and (3) converting human-annotated counterfactual labels to interactions via PID.}
\centering
\begin{tabular}{l|cccc|cccc|cccc|cccc|cccc}
\hline \hline
Task & \multicolumn{4}{c|}{\textsc{VQA 2.0}: image+text} & \multicolumn{4}{c|}{\textsc{CLEVR}: image+text}  & \multicolumn{4}{c|}{\textsc{MOSEI}: video+text} & \multicolumn{4}{c|}{\textsc{UR-FUNNY}: video+text} & \multicolumn{4}{c}{\textsc{MUStARD}: video+text} \\
\hline
Measure & \red & \uone & \utwo & \syn & \red & \uone & \utwo & \syn & \red & \uone & \utwo & \syn & \red & \uone & \utwo & \syn & \red & \uone & \utwo & \syn \\
\hline
Info. decomposition & $0$ & $0$ & $0$ & $\bm{4.97}$ & $0$ & $0$ & $0$ & $\bm{4.76}$ & $\bm{3.37}$ & $\bm{2.07}$ & $1.57$ & $1.51$ & $2.12$ & $\bm{2.86}$ & $1.83$ & $\bm{2.39}$ & $2.12$ & $\bm{2.74}$ & $0.58$ & $\bm{2.08}$ \\
\hline
Partial+PID & $0.12$ & $0.56$ & $1.68$ & $\bm{4.46}$ & $0.66$ & $0.1$ & $0.66$ & $\mathbf{4.10}$ & $\mathbf{0.49}$ & $\mathbf{0.54}$ & $0.11$ & $0.34$ & $0.12$ & $\mathbf{0.20}$ & $0.13$ & $\mathbf{0.20}$ & $0.13$ & $\bm{0.36}$ & $0.01$ & $\mathbf{0.32}$ \\
Counterfactual+PID & $0.32$ & $0.46$ & $0.68$ & $\bm{3.98}$ & $0.10$ & $0$ & $0.48$ & $\bm{4.50}$ & $\bm{0.42}$ & $\bm{0.38}$ & $0.27$ & $0.31$ & $0.04$ & $\bm{0.29}$ & $0.05$ & $0.08$ & $0.09$ & $\bm{0.21}$ & $0.07$ & $\bm{0.16}$\\
\hline \hline
\end{tabular}

\vspace{-1mm}
\label{tab:datasets_fusion}
\end{table*}

We now turn our attention to annotating information decomposition. Referencing the average annotated interactions in Table~\ref{tab:datasets_fusion} with agreement scores in Table~\ref{tab:agreement}, we explain our findings regarding annotation quality and consistency. We also note qualitative feedback from annotators regarding any challenges they faced.
\begin{itemize}[noitemsep,topsep=0pt,nosep,leftmargin=*,parsep=0pt,partopsep=0pt]
    \item \textbf{General observations on interactions, agreement, and confidence}: The annotated interactions align with prior intuitions on these multimodal datasets and do indeed explain the interactions between modalities, such as VQA and CLEVR with significantly high synergy, as well as language being the dominant modality in sentiment, humor, and sarcasm (high $U_1$ values). Overall, the Krippendorff's alpha for inter-annotator agreement in directly annotating the interactions is quite high (roughly $0.5$ for each interaction) and the average confidence scores are also quite high (above $4$ for each interaction), indicating that the human-annotated results are reasonably reliable. 

    \item \textbf{Uniqueness vs synergy in video datasets}: There was some confusion between uniqueness in the language modality and synergy in the video datasets, resulting in cases of low agreement in annotating $U_1$ and $S$: $-0.09$, $-0.07$ for \textsc{MOSEI}, $-0.14$, $-0.03$ for \textsc{UR-FUNNY} and $-0.08$, $-0.04$ for \textsc{MUStARD} respectively. We believe this is due to subjectivity in interpreting whether sentiment, humor, and sarcasm are present in the language only or present only when contextualizing both language and video.
    
    \item \textbf{Information decomposition in non-video datasets}: On non-video datasets, there are cases of disagreement due to the subjective definitions of information decomposition. For example, there was some confusion regarding \textsc{VQA} and \textsc{CLEVR}, where images are the primary source of information that must be selectively filtered by the question. This results in response synergy but information uniqueness. One annotator consistently annotated high visual uniqueness as the dominant interaction, while the other two recognized synergy as the dominant interaction, so the agreement of annotating synergy was low ($-0.04$).
    
    \item \textbf{On presence vs absence of an attribute}: We further investigated the difference between agreement and confidence in the presence or absence of an attribute (e.g., humor or sarcasm). Intuitively, the presence of an attribute is clearer: taking the example of synergy, humans can judge that there is no inference of sarcasm from text only and there is no inference of sarcasm from the visual modality only, but there is sarcasm when both modalities interact together~\cite{castro2019towards}. Indeed, we examined videos that show and do not show an attribute separately and found in general, humans reached higher agreement on annotating attribute-present videos. The agreement of annotating $S$ is $0.13$ when the attribute is present, compared to $-0.10$ when absent.
\end{itemize}
Overall, we find that while annotating information decomposition can perform well, there are some sources of confusion regarding certain interactions and during the absence of an attribute.

\vspace{-1mm}
\subsection{Converting partial and counterfactual labels to information decomposition}

Finally, we present results on converting partial and counterfactual labels into interactions using our information-theoretic method (PID). We report these results in Table~\ref{tab:datasets_fusion} in the rows called Partial+PID and Counterfactual+PID, and note the following:
\begin{itemize}[noitemsep,topsep=0pt,nosep,leftmargin=*,parsep=0pt,partopsep=0pt]

    \item \textbf{Partial+PID vs counterfactual+PID}: In comparing conversions on both partial and counterfactual labels, we find that the final interactions are very consistent with each other: the highest interaction is always the same across the datasets and the relative order of interactions is also maintained.

    \item \textbf{Comparing with directly annotated interactions}: In comparison to the interaction that human annotators rate as the highest, PID also assigns the largest magnitude to the same interaction ($S$ for \textsc{VQA 2.0} and \textsc{CLEVR}, $U_1$ for \textsc{UR-FUNNY} and \textsc{MUStARD}), so there is strong agreement. For \textsc{MOSEI} there is a small difference: both $R$ and $U_1$ are annotated as equally high by humans, while PID estimates $R$ as the highest.

    \item \textbf{Normalized comparison scale}: Observe that the converted results fall into a new scale and range, especially for the \textsc{MOSEI}, \textsc{UR-FUNNY}, and \textsc{MUStARD} video datasets. This is expected since PID conversion inherits the properties of information theory where $R+U_1+U_2+S$ add up to the total information that the two modalities provide about a task, indicating that the three video datasets are more subjective and are harder to predict.

    \item \textbf{Propagation of subjectivity}: On humor and sarcasm, the subjectivity in initial human partial labeling can be propagated when we subsequently apply automatic conversion - after all, we do not expect the automatic conversion to change the relative order apart from estimating interactions in a principled way.
    
\end{itemize}
Therefore, we believe that the conversion method we proposed is a stable method for estimating information decomposition, combining human-in-the-loop labeling of partial labels (which shows high agreement and scales to high-dimensional data) with information-theoretic conversion which enables comparable scales, normalized values, and well-defined distance metrics.

\vspace{-1mm}
\subsection{An overall guideline}

Given these findings, we summarize the following guidelines for quantifying multimodal fusion interactions: 
\begin{itemize}[noitemsep,topsep=0pt,nosep,leftmargin=*,parsep=0pt,partopsep=0pt]
    \item For modalities and tasks that are more objective (e.g., visual question answering), direct annotation of information decomposition is a reliable alternative to conventional methods of partial and counterfactual labeling to study multimodal interactions.
    \item For modalities and tasks that may be subjective (e.g., sarcasm, humor), it is useful to obtain counterfactual labels before using PID conversion to information decomposition values, since counterfactual labeling exhibits higher annotator agreement while PID conversion is a principled method to obtain interactions.
\end{itemize}

\vspace{-1mm}
\section{Conclusion}

Our work aims to quantify various categorizations of multimodal interactions using human annotations. Through a comprehensive study of partial labels, counterfactual labels, and information decomposition, we elucidated several pros and cons of each approach and proposed a hybrid estimator that can convert partial and counterfactual labels to information decomposition interaction estimates. On real-world multimodal fusion tasks, we show that we can estimate interaction values accurately and efficiently which paves the way towards a deeper understanding of these multimodal datasets.

\textbf{Limitations and future work}: The annotation schemes in this work are limited by the subjectivity of the modalities and task. Automatic conversion of partial labels to information decomposition requires the label space to be small and discrete (i.e., classification), and does not yet extend to regression or text answers unless approximate discretization is first performed. Future work can also scale up human annotations to more datapoints and fusion tasks, and ask annotators to provide their explanations for ratings that have low agreement. Finally, we are aware of challenges in evaluating interaction estimation and emphasize that they should be interpreted as a relative sense of which interaction is most important and a guideline to inspire model selection and design.

\vspace{-1mm}
\section*{Acknowledgements}

This material is based upon work partially supported by Meta, National Science Foundation awards 1722822 and 1750439, and National Institutes of Health awards R01MH125740, R01MH132225, R01MH096951 and R21MH130767.
PPL is partially supported by a Facebook PhD Fellowship and a Carnegie Mellon University's Center for Machine Learning and Health Fellowship.
RS is supported in part by ONR N000141812861, ONR N000142312368 and DARPA/AFRL FA87502321015.
Any opinions, findings, conclusions, or recommendations expressed in this material are those of the author(s) and do not necessarily reflect the views of the NSF, NIH, Meta, CMLH, ONR, DARPA, or AFRL, and no official endorsement should be inferred. We are grateful to the anonymous reviewers for their valuable feedback. Finally, we would also like to acknowledge NVIDIA’s GPU support.

\clearpage

\bibliographystyle{ACM-Reference-Format}
\balance
\bibliography{sample-bibliography}

\clearpage

\appendix

\section{Human Annotation Details}
\label{app:details}

Participation in all annotations was fully voluntary and we obtained consent from all participants prior to annotations. The authors manually took anonymous notes on all results and feedback in such a manner that the identities of annotators cannot readily be ascertained directly or through identifiers linked to the subjects. 
Participants were not the authors nor in the same research groups as the authors, but they all hold or are working towards a graduate degree in a STEM field and have knowledge of machine learning. None of the participants knew about this project before their session and each participant only interacted with the setting they were involved in.

We sample $50$ datapoints from each of the $5$ datasets in Table~\ref{tab:setup} and give them to a total of $18$ different annotators:
\begin{itemize}[noitemsep,topsep=0pt,nosep,leftmargin=*,parsep=0pt,partopsep=0pt]
	\item 3 annotators for direct annotation of interactions,
	\item 3 annotators for partial labeling of $y_1$, $y_2$, and $y_{12}$,
	\item 3 annotators for counterfactual, labeling $y_1$ first then $y_{1+2}$,
	\item 3 annotators for counterfactual, labeling $y_2$ first then $y_{2+1}$.
\end{itemize}
All annotations were performed via google spreadsheets.

\subsection{Annotating partial labels}
\label{app:indirect1}
We asked 3 annotators to predict the partial labels in a randomized setting. For each annotator, we asked them to annotate $y_1$ then $y_2$ given only modality 1 or 2 respectively, and finally $y$ given both modalities. This completion order is designed on purpose to minimize possible memorization of the data so that the annotators can provide completely independent unimodal and multimodal predictions on the label. When annotating the visual modality of the video datasets, we explicitly require the annotators to mute the audio and predict the partial labels based only on the video frames. After that, all annotators are asked to provide a confidence score on a scale of $0$ (no confidence) to $5$ (high confidence) about their annotations. The confidence scale is applied to all annotation settings below. We aggregated annotator $A$'s $y_1$ response, annotator $B$'s $y_2$ response, and annotator $C$'s $y$ response as one set of complete partial labels. Similarly, we collected $B$'s $y_1$, $C$'s $y_2$, and $A$'s $y$ as the second set, $C$'s $y_1$, $A$'s $y_2$, and $B$'s $y$ as the third set.

\subsection{Annotating counterfactual labels}
\label{app:indirect2}
We asked 6 annotators to predict the counterfactual labels in this setting. For each group of 2 annotators, we asked the first annotator to annotate partial labels $y_1$ given only the first modality and provide confidence scores, then presented them with the other modality and asked for their new predictions $y_{1+2}$ and corresponding confidence ratings. We asked the second annotator to predict $y_2$ similarly with only the second modality and then $y_{2+1}$ with both modalities presented.

\subsection{Annotating information decomposition}
\label{app:direct}

We asked 3 annotators to directly annotate the information decomposition values. Given both modalities, each annotator is asked to provide a rating on a scale of $0$ (none at all) to $5$ (large extent) for the following questions that correspond to $R$, $U_1$, $U_2$, and $S$ respectively:
\begin{enumerate}
    \item The extent to which both modalities enable them to make the same predictions about the task;
    \item The extent to which modality 1 enables them to make a prediction that they would not if using modality 2;
    \item The extent to which modality 2 enables them to make a prediction that they would not if using modality 1;
    \item The extent to which both modalities enable them to make a prediction that they would not if using either modality individually.
\end{enumerate}
Finally, they are asked to rate their confidence for each rating they provided, on a scale of $0$ (no confidence) to $5$ (high confidence).

\subsection{Video and audio presentation}

Annotators were provided with the full video link which opens up in a separate video player. They asked to either annotate based on all modalities in the video (i.e., video + audio), or asked to mute the videos themselves when annotating based on vision only, or are not provided the video at all when annotating based only on the transcript. We did not completely remove the audio from videos because in all tasks, annotators have to use video only (with mute), followed by audio+transcripts, and finally with all modalities (video+audio+transcripts). In the counterfactual setting they may see video only (with mute), before playing the entire video with audio. Hence, we instructed the annotators to mute the videos for video-only prediction, and unmute the video for predictions that involve audio. We specifically checked with the annotators and they strictly followed these guidelines.

\subsection{Label space for QA tasks}

For VQA and CLEVR datasets, annotators were requested to write the answer themselves. For CLEVR the answer is always yes/no. For VQA we let the users write their own answer, but we post-hoc modify these answers to define a similarity with the final answer $y$: whether $y_1$ or $y_2$ are the same as multimodal $y$, or otherwise different. This binary distance function is sufficient to distinguish different interactions.

\end{document}